\pdfoutput=1
\documentclass[11pt]{article}

\usepackage[]{ACL2023}
\usepackage{graphicx}
\usepackage{wrapfig}
\usepackage{booktabs} 
\usepackage{xspace}
\usepackage{multirow,multicol}
\usepackage{enumitem}
 
\usepackage{algorithm}
\usepackage{amssymb, latexsym, amsmath,bm,lipsum}
\usepackage{subcaption}
\usepackage{algpseudocode}
\usepackage[marginal]{footmisc}
\usepackage[utf8]{inputenc}
\usepackage{amsthm}
\usepackage[switch]{lineno}
\usepackage{xcolor}
\usepackage{enumitem}
\usepackage{longtable} % Allows tables to span multiple pages 
\usepackage{threeparttable}

\usepackage{hyperref}
\usepackage{url}

\usepackage{times}
\usepackage{latexsym}

\usepackage[T1]{fontenc}
\usepackage[utf8]{inputenc}

\usepackage{microtype}

\usepackage{inconsolata}
\usepackage{CJKutf8}
\usepackage{tabularx}

\title{ZiGong 1.0: A Large Language Model for Financial Credit}

\author{Yu Lei\textmd{,} Zixuan Wang\textmd{,}  Chu Liu \textmd{and} Tongyao Wang \\
Didi International Business
\vspace{1mm} \\
  \texttt{ leiyu0210@gmail.com},\ \ \ 
   \texttt{ \{maxwangzixuan, liuchu, wangtongyao\}@didiglobal.com}
}

\begin{document}
\maketitle
\begin{abstract}
Large Language Models (LLMs) have proven their effectiveness in a variety of general Natural Language Processing (NLP) tasks. However, their performance in financial credit assessment tasks has yet to reach its full potential, partly because these tasks require specific financial credit expertise. To address this challenge, we propose the ZiGong (\begin{CJK*}{UTF8}{gbsn}子贡\end{CJK*}) model, based on Mistral, which employs multi-task supervised fine-tuning. Furthermore, to address the issue of model hallucination in financial scenarios, we propose a novel data pruning method. Specifically, we employ an agent model to assign scores to training samples, and then integrate the pruned samples with the original data for model training. This approach effectively mitigates hallucinations in large models by refining the training data, ensuring higher reliability in downstream applications. Experimental results demonstrate that our method significantly improves the model’s robustness and accuracy in real-world financial scenarios.
\end{abstract}

\section{Introduction}

In the ever-evolving landscape of finance, credit risk assessment remains a cornerstone for maintaining economic stability and fostering growth. With the financial industry increasingly embracing digital transformation, the need for sophisticated and adaptable credit evaluation mechanisms has never been more critical~\cite{lei2024finlangnet}. Traditional credit assessment methods, often reliant on rule-based systems and specific domain expertise, struggle to keep pace with the dynamic market environment and the vast array of available data sources. These limitations underscore the necessity for more flexible, generalist approaches capable of leveraging diverse datasets and providing robust predictive insights across multiple financial tasks.

The advent of Large Language Models (LLMs) presents a transformative opportunity to revolutionize credit risk evaluation. LLMs~\cite{zhao2024revolutionizing}, with their ability to process and understand complex natural language inputs, can integrate vast amounts of historical and real-time data, capturing nuanced patterns and trends that were previously unattainable. By employing techniques such as multi-task learning and few-shot generalization, these models can seamlessly adapt to various credit-related tasks, from predicting loan defaults to identifying potential fraud, all while drawing on a comprehensive knowledge base spanning different financial domains.

Moreover, Large Language Models (LLMs) in the financial credit domain still suffer from issues such as hallucinations~\cite{huang2024survey} and knowledge forgetting~\cite{luo2023empirical}. To address these challenges, we propose a novel data pruning approach for constructing high-quality instruct data. Specifically, we utilize an agent model (a domain-specific lightweight model) to score training samples and select the Top-K samples for mixed training. Building upon our optimized version of TracInCP~\cite{pruthi2020estimating}, we further incorporate the sequential temporal characteristics of financial data and propose TracSeq, a data distillation method designed to improve training stability and reduce hallucination effects. By leveraging the time-dependent nature of financial behavior data, TracSeq ensures that the model better retains critical knowledge while minimizing noise, leading to enhanced reliability in downstream financial applications.

The main contributions are summarized as follows:
\begin{enumerate}[topsep=-2pt, partopsep=0pt, parsep=0pt, itemsep=2pt]
\item We propose TracSeq, which integrates temporal dependencies in financial data to enhance data distillation, preserve long-term knowledge, and reduce catastrophic forgetting.

\item Our hybrid training approach combines Top-K selected instruct data with the original dataset, improving model robustness, mitigating hallucinations, and enhancing generalization in real-world financial applications.

\item This method has been successfully deployed in our Behavior Card service, which supports the operational model in the loan process. Notably, ZiGong achieves superior performance on the benchmark.
\end{enumerate}
\section{Related Work}
\subsection{Large Language Models}
Large Language Models (LLMs) have significantly evolved over recent years, demonstrating remarkable capabilities across a variety of natural language processing tasks. Notable models such as DeepSeek V3~\cite{liu2024deepseek}, GPT-4~\cite{achiam2023gpt}, Llama ~\cite{touvron2023llama}, and more recently, Mistral 7B~\cite{jiang2023mistral}, have set new benchmarks in terms of language understanding and generation capabilities. Mistral 7B, for example, presents a sophisticated architecture that balances model complexity with computational efficiency, supporting more nuanced text generation and comprehension tasks. These models are pretrained on extensive datasets, allowing them to generalize well across diverse linguistic contexts and effectively handle tasks such as translation, summarization, and question answering.

\subsection{Large Language Models in Financial Domain}
The application of large language models within the financial domain is an emerging area of interest, driven by the potential to transform traditional financial analysis and decision-making processes. LLMs are increasingly being used to enhance financial forecasting, sentiment analysis, and risk management by leveraging their ability to process and analyze large volumes of textual data from financial reports, news articles, and social media. For instance, FinBERT, a specialized variant of BERT, has been fine-tuned for finance-specific tasks and shows improved performance in sentiment analysis of news headlines and economic reports. XuanYuan ~\cite{zhang2023xuanyuan}.

Recent studies have explored the potential of using LLMs for predictive analytics in finance, such as predicting stock movements and credit scoring. These models are designed to discern financial insights from unstructured data, complementing traditional quantitative models. Moreover, LLMs offer an advantage in terms of scalability and adaptability, making them suitable for real-time analysis and decision support in fluctuating market conditions.

Furthermore, the integration of LLMs in financial services raises questions of ethical AI and fairness, particularly concerning biases inherent in training data that could affect financial decision-making. Therefore, ongoing research focuses on developing techniques for bias mitigation and ethical deployment of LLMs to ensure fair, transparent, and accountable use in financial applications.

\section{ZiGong Model}
\begin{figure*}[!t]
\centering
\includegraphics[width=\linewidth]{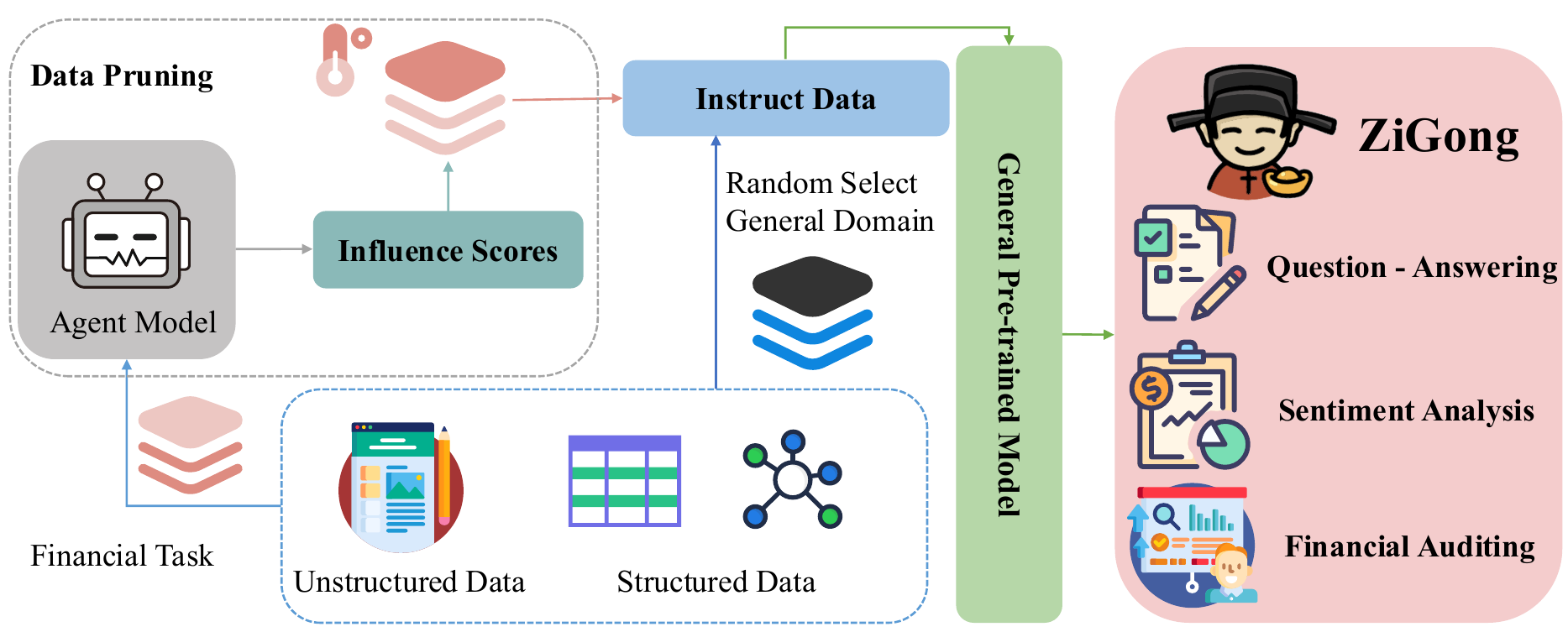}
\caption{Illustrates the entire workflow of ZiGong.}
\label{fig:ZiGong}
\vspace{-5pt}
\end{figure*}

In this section, we will introduce the training process of our ZiGong (\begin{CJK*}{UTF8}{gbsn}子贡\end{CJK*}) model.
Figure \ref{fig:ZiGong} illustrates the entire workflow of ZiGong. The process begins by applying data pruning and random selection strategies to select samples from the dataset, which are then used to construct instruct data. Based on the Mistral 7B foundation model, various downstream tasks, including QA (Question - Answering), Sentiment Analysis, and Financial Auditing, are developed to build the ZiGong financial credit large model.

\subsection{Data Pruning}

Constructing high-quality data is crucial for LLMs, and many studies have adopted data distillation techniques to optimize training datasets. DEALRec~\cite{lin2024data} leverages an influence score to evaluate each data point’s contribution, while \cite{li2023one} introduces a PPL (perplexity)-based metric to quantify data impact on downstream tasks, capturing its intrinsic relevance. Additionally, LESS~\cite{xia2024less} and LIMA~\cite{zhou2024lima} apply data distillation to refine the fine-tuning process of LLMs. These approaches highlight the importance of selecting high-value data to enhance training efficiency and model generalization.

TracInCP~\cite{pruthi2020estimating} is an effective method for measuring training data influence by leveraging gradient similarity at stored checkpoints. However, its original formulation is primarily designed for image data, where each sample is treated independently without considering temporal dependencies. This assumption does not hold in sequential behavior data, where user actions evolve over time and past behaviors impact future predictions. Simply treating different time periods as separate, independent samples leads to information loss, as it ignores the inherent time dependencies within each user’s behavior sequence.

To address this issue, we propose an improved version of TracInCP, denoted as \textbf{TracSeq}, which incorporates temporal dependencies in training data influence estimation. Instead of treating different time periods of the same user as independent instances, we introduce a \textit{time decay factor} to model the diminishing influence of past behaviors over time. The modified influence estimation formula is as follows:

\begin{equation}
\text{TracSeq}= \sum_{i=1}^{k} \gamma^{T - t_i} \eta_i \nabla \ell(w_{t_i}, z_t) \cdot \nabla \ell(w_{t_i}, z'_T)
\label{eq:tracin-seq}
\end{equation}
where
    \( z_t \) represents the training data corresponding to a user's behavior at time \( t \).
    \( z'_T \) is the test data corresponding to the user's behavior at the current time \( T \).
    \( w_{t_i} \) denotes the model parameters stored at checkpoint \( t_i \).
    \( \nabla \ell(w_{t_i}, z_t) \) and \( \nabla \ell(w_{t_i}, z'_T) \) represent the loss function gradients of training and test samples at checkpoint \( t_i \), respectively.
    \( \eta_i \) is the step size used between checkpoint updates.
    \( \gamma^{T - t_i} \) is a \textbf{time decay factor}, where \( \gamma \in (0,1] \) controls the rate at which past influences diminish over time.

To construct the dataset \( D \), we select the \textbf{Top-\( k \) most influential} training samples based on the \textbf{TracSeq} score. Based on this influence metric, we define the selected dataset \( D \) as:

\begin{equation}
D = \{ z_t \mid z_t \in \text{Top-} k \text{  TracSeq}(z_t) \}
\end{equation}

The dataset \( D \) consists of the most valuable training samples, which are expected to improve model generalization. The \textbf{time decay factor} \( \gamma^{T - t_i} \) ensures that more recent samples receive higher weights, accounting for potential data distribution shifts over time. By filtering out less impactful samples, this method improves training efficiency and refines the training set for fine-tuning large language models.

\subsection{Financial Credit Instruction Data}
In this training process, 70\% of the samples are randomly selected from the entire dataset, while the remaining 30\% are high-influence samples filtered through data pruning. These selected samples are then used to construct the instruct data according to the tasks outlined below. All task types considered and their corresponding prompt templates are presented in Table~\ref{tab:eval-method}.

The German and Australia datasets focus on Credit Scoring, a crucial process in financial institutions for assessing a borrower's likelihood of repaying debt. This involves analyzing an individual's financial information provided during a credit application. The analysis helps lenders determine loan eligibility, interest rates, and repayment terms. Closely related to credit scoring is Fraud Detection, which aims to identify whether an online loan application is legitimate or fraudulent. This process is essential for maintaining the integrity of financial systems and protecting institutions and customers from financial losses. Two key datasets used for this task are Credit Card Fraud and ccFraud. For the Travel Insurance dataset, the focus is on Claim Analysis, a critical task for insurance companies to detect fraudulent claims. Fraudulent claims involve false or deceptive information intended to receive undeserved payouts, while legitimate claims are valid and result from genuine losses covered by insurance policies. Distinguishing between these two is crucial to preventing financial losses due to fraud and ensuring that only valid claims are processed.

The datasets mentioned above are used for Discriminative tasks, whereas real-world often require Generative tasks. In practical applications, data is collected through QA-based questioning, where a user’s basic attributes (such as gender, age, education level, partial residential address, past job earnings, transaction history, events, and locations) are analyzed to construct a financial profile. Additionally, details like mobile phone brand, model, price, and purchase year are utilized to predict the user’s income through regression-based models.

\begin{table}[t]
\small
\centering
\begin{tabular}{@{}p{0.3\linewidth}@{} p{0.7\linewidth}@{}}
\toprule
Task & Template/Example \\ \midrule
\textbf{Discriminative} \\ \cmidrule{1-1}
Sentiment Analysis & \texttt{\{sentence\}\newline Question:~what is the sentiment?\newline Answer:~\{good/neutral/bad\}} \\ 
Classification & \texttt{\{sentence\}\newline Question:~\{question\}?\newline Answer:~\{Yes/No\}} \\
\textbf{Generative} \\ \cmidrule{1-1}
QA & \texttt{\{context\}\newline Question:~\{question\}?\newline Answer:~\{answer\}} \\
\bottomrule
\end{tabular}
\caption{Template for the different tasks we evaluate in the financial credit.}
\label{tab:eval-method}
\end{table}

\begin{table*}[h]
\centering
\vspace{-1mm}
\caption{The performance of LLMs and the SOTA expert system models on our benchmark. We use bold to indicate the best and underline to indicate the second-best. For Miss, where smaller is better, for other metrics, larger is better.}
\label{tab:bench}
 \vspace{-3mm}
\resizebox{0.999\textwidth}{!}{%
    \begin{threeparttable}
    \begin{tabular}{cccccccccccc}
    \toprule
    Dataset  & Metrics &  ChatGPT & GPT4  & Bloomz & Vicuna & Llama1 & Llama2 & Llama2-chat & FinMA & CALM & ZiGong \\
    \midrule
    \multirow{3}[2]{*}{German}  & Acc   & 0.440 & 0.545 & 0.315 & \underline{0.590} & \textbf{0.660} & \textbf{0.660} & 0.475  & 0.170 & 0.565 & \underline{0.590}\\
          &       F1    & 0.410 & 0.513 & 0.197 & 0.505 & 0.173 & 0.173 & 0.468 & 0.170  & \underline{0.535} & \textbf{0.587} \\
          &        Miss   & 0.000 & 0.000 & 0.110 & 0.000 & 0.000 & 0.000 & 0.000 & 0.110 & 0.000 & 0.000 \\
    \midrule
    \multirow{3}[2]{*}{Australia}  & Acc   & 0.425 & \underline{0.748} & 0.568 & 0.489 & 0.432 & 0.432 & 0.432 & 0.410 & 0.518 & \textbf{0.779} \\
          &       F1    &  0.257 & \underline{0.746} & 0.412 & 0.513 & 0.412 & 0.412 & 0.260  & 0.410 & 0.492 & \textbf{0.777}\\
          &        Miss     & 0.000 & 0.000 & 0.000 & 0.000 & 0.000 & 0.000 & 0.000 & 0.806 & 0.000 & 0.014 \\
    \midrule
   
    \multirow{3}[2]{*}{Credit Card Fraud}  & Acc   & \underline{0.998} & 0.810 & 0.001 & \textbf{0.999} & 0.823 & \textbf{0.999} & -     & 0.003 & 0.947 & \underline{0.998}\\
          &        F1    & \underline{0.998} & 0.878 & 0.000 & \underline{0.998} & 0.902 & \underline{0.998} & -     & 0.004 & 0.971 & \textbf{0.999}\\
          &        Miss       & 0.000 & 0.110 & 0.000 & 0.000 & 0.176 & 0.000 & 1.000  & 0.000 & 0.000 & 0.031\\
    \midrule
    \multirow{3}[2]{*}{ccFraud} & Acc   & 0.173 & 0.580 & 0.059 & 0.608 & \underline{0.941} & \underline{0.941} & 0.914  & 0.060 & 0.514 & \textbf{0.987} \\
          &       F1    &  0.214 & 0.587 & 0.007 & 0.651 & 0.007 & 0.007 & 0.437  & -0.006 & 0.627 &\textbf{0.982}\\
          &       Miss     & 0.000 & 0.210 & 0.000 & 0.000 & 0.000 & 0.000 & 0.000 & 0.891 & 0.000 & 0.000 \\
    \midrule
    
    \multirow{3}[2]{*}{Travel Insurance}  & Acc    & \textbf{0.981} & 0.835 & 0.015 & 0.015 & 0.000 & 0.015 & 0.665  & 0.002 & \underline{0.929} & 0.884 \\
          &       F1     & \underline{0.975} & 0.897 & 0.000 & 0.130 & 0.001 & \textbf{0.978} & 0.787  &0.001 & 0.950 & 0.927 \\
          &       Miss     & 0.000 & 0.000 & 0.000 & 0.000 & 0.999 & 0.000 & 0.000 & 0.000 & 0.000  & 0.064\\
    \bottomrule
    \end{tabular}%%%
    
 \begin{tablenotes}
   \item[-] The related studies balance the data for the test set, and the values are for reference only.
    \end{tablenotes}
    \end{threeparttable}
}
 \vspace{-6mm}
\end{table*}

\section{Experiment}
We conducted our experiments on the benchmark provided by CALM~\cite{feng2023empowering}, which includes 4 tasks: credit scoring, fraud detection, financial distress identification, and claim analysis. Among these datasets, we evaluated our model's performance comprehensively.
When finetuning our LLM (ZiGong), we use the LORA ~\cite{hu2021lora} to reduce the computation cost. Configuration Details of ZiGong are presented in Table~\ref{tab:zigong-config}.
\begin{table*}[h]
\centering
\caption{Configuration Details of ZiGong Model (Mistral 7B Fine-tuned)}
\label{tab:zigong-config}
\begin{tabular}{>{\bfseries}llp{7cm}}
\toprule
\multicolumn{1}{l}{\textbf{Category}} & \textbf{Parameter} & \textbf{Configuration} \\
\midrule

\multicolumn{3}{l}{\textbf{Base Information}} \\
\midrule
& Model Name & ZiGong \\
& Base Model & Mistral 7B \\
& Fine-tuning Method & LoRA (Low-Rank Adaptation) \\
& Task Type & Text Generation \& Classification \\
& Context Length & 4096 tokens (unchanged) \\
\midrule

\multicolumn{3}{l}{\textbf{Architecture}} \\
\midrule
& Parameters & 7B (base) + LoRA Adapters \\
& Hidden Dimension & 4096 \\
& Attention Heads & 32 \\
& Layers & 32 \\
& Activation Function & SiLU (unchanged) \\
\midrule

\multicolumn{3}{l}{\textbf{Training Configuration}} \\
\midrule
& Learning Rate & $1\times10^{-5}$ -- $3\times10^{-5}$ \\
& Batch Size & 32 (with gradient accumulation: 4) \\
& Optimizer & AdamW ($\beta_1=0.9$, $\beta_2=0.999$) \\
& LR Schedule & Cosine Decay \\
& Max Sequence Length & 4096 tokens \\
& Training Steps & [Insert actual steps] \\
& LoRA Rank & 8 \\
& LoRA Alpha & 16 \\
& Target Modules & \{query, key, value\} \\
\midrule

\end{tabular}
\end{table*}

\section{Results}
Table \ref{tab:bench}  shows the results of the benchmark with the LLMs. The results demonstrate that ZiGong consistently achieves strong performance across multiple datasets, outperforming or matching other models, particularly excelling in tasks like "Australia" and "Credit Card Fraud." This highlights ZiGong’s effectiveness in financial credit applications.

\begin{figure}[!t]
\centering
\includegraphics[width=\linewidth]{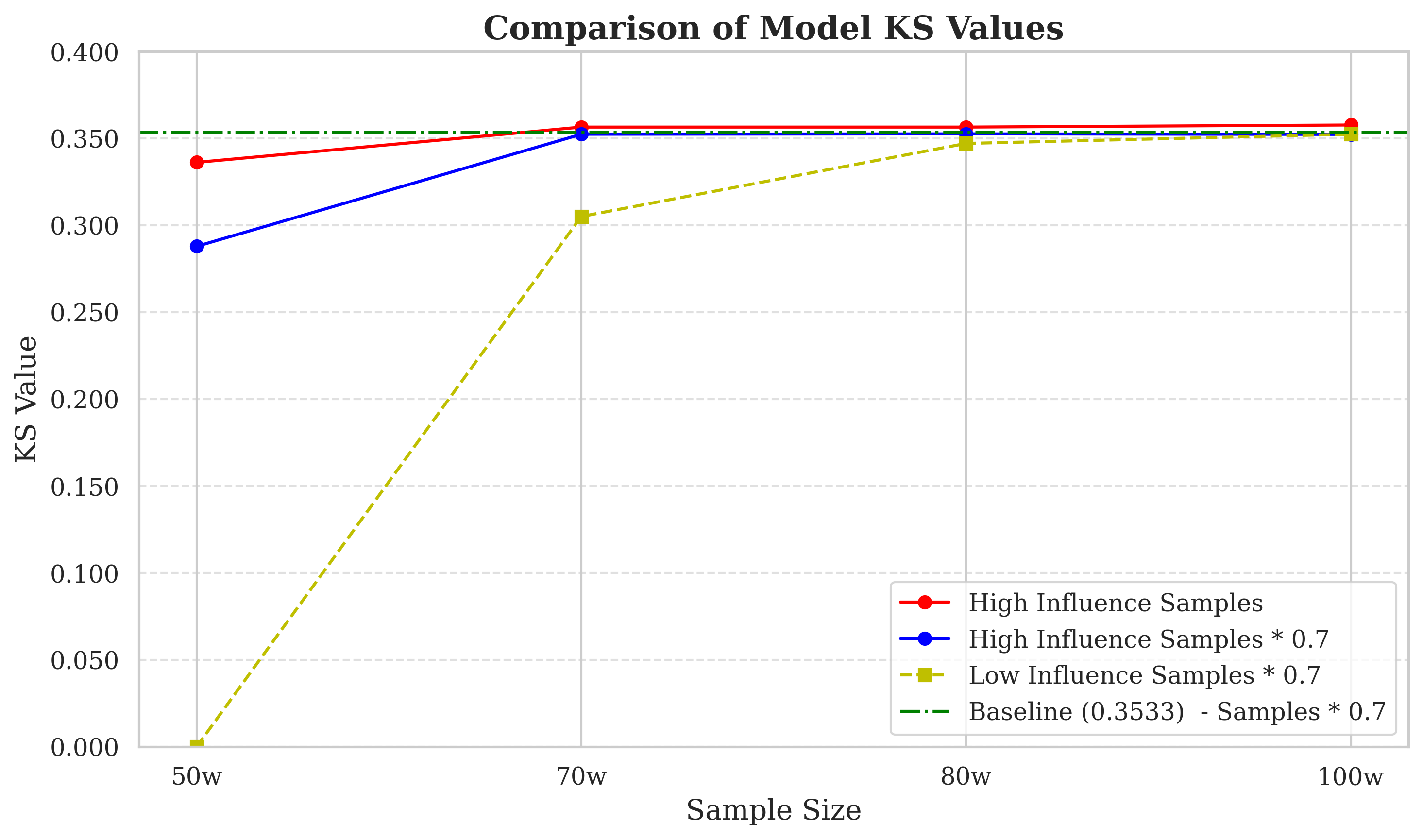}
\caption{Illustrates the impact of data pruning on model performance by comparing results across different sample sizes.}
\label{fig:data_pruning}
\vspace{-5pt}
\end{figure}

Figure \ref{fig:data_pruning} illustrates the impact of data pruning on model performance by comparing results across different sample sizes. Specifically, the figure demonstrates the performance disparity between selecting high-influence samples and low-influence samples. The findings reveal that utilizing only half of the high-influence samples can achieve superior performance compared to using the entire original dataset. This highlights the efficiency of data pruning in retaining the most impactful data points. Additionally, the KS (Kolmogorov-Smirnov) metric, which is a critical measure in the financial risk control domain, is employed to evaluate the model's effectiveness. The improved KS scores further confirm that focusing on high-influence samples enhances the model's capability in financial risk assessment.

\section{Conclusion}
LLMs in the financial credit domain continue to face significant challenges, including hallucinations and knowledge forgetting. To address these issues, we introduce a novel data pruning approach that constructs high-quality instructional data by leveraging a domain-specific lightweight agent model to score and select the Top-K training samples for mixed training. Building on our optimized version of TracInCP, we present TracSeq, a data distillation method that incorporates the sequential temporal characteristics of financial data. This integration enhances training stability and reduces hallucination effects by utilizing the time-dependent nature of financial behavior data, ensuring the model retains critical knowledge while minimizing noise. Our hybrid training strategy, which combines Top-K selected instructional data with the original dataset, significantly improves model robustness, mitigates hallucinations, and enhances generalization in real-world financial applications. Furthermore, the successful deployment of this method in our Behavior Card service, supporting the operational model in the loan process, demonstrates its practical efficacy, with ZiGong achieving superior performance on benchmark evaluations. These advancements collectively enhance the reliability and effectiveness of LLMs in downstream financial applications.

\section{Discussion}
The application of financial credit large models extends beyond personal and corporate credit assessments to areas such as fraud detection, customer relationship management, and marketing. As the technology matures, these models will increasingly unlock potential in improving credit decision efficiency, reducing financial risk, and advancing financial inclusion. In the future, financial credit large models will not only enhance risk management levels for financial institutions but will also play a pivotal role in driving the intelligent transformation of the entire financial industry. 
\section{Ethics Statement}
The ZiGong is primarily dedicated to research in the financial credit sector and is not intended to offer investment or financial advice. The information and data used by this model are sourced from publicly available financial datasets and knowledge graphs, as well as proprietary internal data. It is important to note that the accuracy of responses generated by large language models cannot be guaranteed, and the financial information utilized therein should not be construed as a substitute for professional investment advice. Before making any financial decisions, it is strongly recommended to consult a qualified financial advisor or professional.

% Entries for the entire Anthology, followed by custom entries
\bibliography{custom}
\bibliographystyle{acl_natbib}

\end{document}